\documentclass[conference]{IEEEtran}
\IEEEoverridecommandlockouts
\usepackage{graphicx} 
\usepackage{amsmath}
\usepackage{amssymb}  
\usepackage{amsfonts}
\usepackage{algorithmic}
\usepackage{array}
\usepackage{textcomp}
\usepackage{stfloats}
\usepackage{url}
\usepackage{verbatim}
\usepackage{graphicx}
\usepackage{caption}
\usepackage{subcaption}
\usepackage[ruled, vlined, linesnumbered]{algorithm2e}
\usepackage{cite}
\usepackage{amssymb,amsfonts,bm}
\usepackage{amsmath,tikz}
\usepackage{color}
\usepackage{mathtools}
\usepackage{rotating}
\usepackage{blkarray}
\usepackage{physics}
\usetikzlibrary{arrows}

\usepackage{amsthm}
\usepackage{comment}
\usepackage{xcolor}  
\definecolor{mygreen}{RGB}{0,150,0}

\usepackage[
    colorlinks=true,
    linkcolor=black,     
    citecolor=black,     
    filecolor=black,
    urlcolor=mygreen     
]{hyperref}

\usepackage{geometry}
\begin{document}

\title{\LARGE \bf
AV-PedAware: Self-Supervised Audio-Visual Fusion for Dynamic Pedestrian Awareness\\
\author{Yizhuo Yang*,~Shenghai Yuan*,~Muqing Cao,~Jianfei Yang,~Lihua Xie,~\textit{Fellow,~IEEE}
\thanks{This research is supported by the National Research Foundation, Singapore, under its Medium-Sized Center for Advanced Robotics Technology Innovation (CARTIN) and under project WP5 within the Delta-NTU Corporate Lab with funding support from A*STAR under its IAF-ICP program (Grant no: I2201E0013) and Delta Electronics Inc. }
\thanks{$\ast$ Equal contribution. Yizhuo Yang, Shenghai Yuan, Muqing Cao, Jianfei Yang and Lihua Xie are with School of Electrical and Electronic Engineering, Nanyang Technological University, 50 Nanyang Avenue, Singapore 639798, 
   {\tt\small yang0670@e.ntu.edu.sg, shyuan@ntu.edu.sg, mqcao@ntu.edu.sg, jianfei.yang@ntu.edu.sg, elhxie@ntu.edu.sg.}}%

}

}

\maketitle
\IEEEpubid{\begin{minipage}{\textwidth}\ \\[55pt]
\centering
\fbox{%
\parbox{\dimexpr\textwidth-2\fboxsep-2\fboxrule}{%
\centering
\footnotesize
This work has been accepted for publication at the 2023 IEEE/RSJ International Conference on Intelligent Robots and Systems (IROS) © 2023 IEEE.
Personal use of this material is permitted. However, permission must be obtained from IEEE for all other uses,
including reprinting or redistribution, creating derivative works, or reuse of any copyrighted components
of this work in other media.
}%
}
\end{minipage}}

\begin{abstract}
In this study, we introduce AV-PedAware, a self-supervised audio-visual fusion system designed to improve dynamic pedestrian awareness for robotics applications. Pedestrian awareness is a critical requirement in many robotics applications. However, traditional approaches that rely on cameras and LIDARs to cover multiple views can be expensive and susceptible to issues such as changes in illumination, occlusion, and weather conditions. Our proposed solution replicates human perception for 3D pedestrian detection using low-cost audio and visual fusion. This study represents the first attempt to employ audio-visual fusion to monitor footstep sounds for the purpose of predicting the movements of pedestrians in the vicinity. The system is trained through self-supervised learning based on LIDAR-generated labels, making it a cost-effective alternative to LIDAR-based pedestrian awareness. AV-PedAware achieves comparable results to LIDAR-based systems at a fraction of the cost. By utilizing an attention mechanism, it can handle dynamic lighting and occlusions, overcoming the limitations of traditional LIDAR and camera-based systems. To evaluate our approach's effectiveness, we collected a new multimodal pedestrian detection dataset and conducted experiments that demonstrate the system's ability to provide reliable 3D detection results using only audio and visual data, even in extreme visual conditions. We will make our collected dataset and source code available online for the community to encourage further development in the field of robotics perception systems. Code \url{https://github.com/yizhuoyang/AV-PedAware}.
\end{abstract}


\section{Introduction}

Pedestrian awareness is a crucial factor for many autonomous systems, such as delivery robots \cite{rezeck2021cooperative}, COBOTs \cite{el2017design}, and  manipulators arms \cite{razjigaev2021snakeraven}, in ensuring efficient planning and safety in complex and dynamic environments \cite{ji2022robust}. The primary approach to achieving pedestrian awareness has been using exteroceptive sensors such as cameras, 3D-LIDARs \cite{yuan2021survey}, and UWB \cite{nguyen2019single}, which provides high-resolution environmental information. However, these sensors have certain limitations, such as vulnerability to lighting conditions and occlusion, high costs, and the need for additional operator hardware.

In recent years, there has been a growing interest in utilizing audio sensors as a supplementary modality for robot navigation, as evidenced by numerous research studies \cite{valverde2021there,gan2019self,vasudevan2020semantic,masuyama2020self,jain2021vinet,he2018deep,li2022multi,yun2021pano}. Audio sensors can offer valuable environmental information that may not be available through visual and LIDAR, such as sound source direction and distance, and acoustic properties of the surroundings. Additionally, audio sensors are less susceptible to occlusion and can function in low-light or complete darkness. Thus, integrating audio sensors with visual sensors can provide a more complete and robust perception of the environment, which can improve a robot's ability to navigate effectively in challenging environments, and offer a more affordable solution as well.

\begin{figure}[t]
  \centering
   \includegraphics[width=1\linewidth]{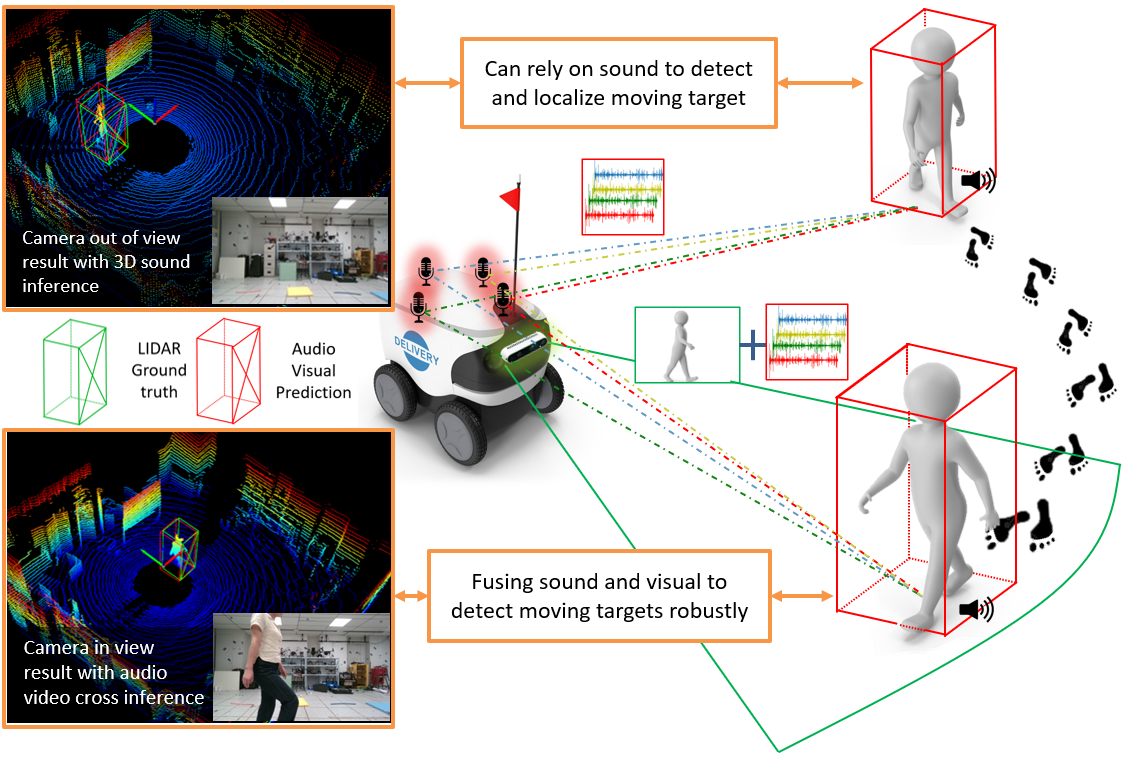}
   \caption{Our proposed method fully exploits the complementary information between audio and visual modalities, enabling 360-degree pedestrian detection.}
   \label{figmoti}
\end{figure}

A promising approach for enhancing pedestrian awareness using audio modality is through self-supervised learning \cite{valverde2021there,gan2019self,masuyama2020self}. Thanks to the co-occurrence among various modalities, the audio-visual object detection model can be trained in a self-supervised mode without the need for any labels. Specifically, the 3D position of a person at a certain moment has natural correspondences in both the camera view and the associated sound. We aim to exploit the spatial and temporal relationships between different modalities cues to enable the model to associate sounds and images with the 3D location of the pedestrians.

In this study, we introduce a new self-supervised cross-model distillation network for 3D pedestrian detection. The proposed network employs a teacher-student network structure, where the teacher network is a well-trained 3D detection model used to generate pseudo-labels for supervising the learning of the audio-visual network. During inference, the network only requires audio and visual signals as inputs to obtain the 3D position information of pedestrians. We evaluate the performance of the proposed approach on a recently collected multimodal dataset, and the experimental results demonstrate the effectiveness of our model in 3D pedestrian detection.

To summarize, this paper presents three key contributions:
\begin{itemize}
\item Our initial contribution is the presentation of a novel network that incorporates both audio and visual data to detect 3D objects. Our network is particularly noteworthy because it can accurately identify pedestrians within a complete 360-degree field of view, including those who are out of camera view and in the presence of challenging visual conditions.
\item Our second contribution is the proposal of a network that operates in a self-supervised learning paradigm, thereby reducing the need for labeled data during the training process. By leveraging the natural correspondence between different modalities, our approach achieves high levels of accuracy while minimizing the need for manual annotation. As a result, our proposed method is both efficient and effective, offering significant advantages over traditional supervised learning methods.

\item Our final contribution is the introduction of a multimodal dataset that includes point cloud, RGB image, and audio data of pedestrians. This dataset is a crucial resource for researchers investigating sound-based object localization, providing valuable data for future studies in this area. Importantly, to the best of our knowledge, this is the first dataset that incorporates audio data in combination with traditional modalities in this manner.

\end{itemize}

\section{Related work}
Previous research in audio has largely centered on natural language processing (NLP) and its use as a means of communication with robots \cite{pramanick2019enabling,hara2004robust}. However, more recently, a growing number of researchers have been exploring the potential of multimodal approaches that integrate audio and visual data to detect \cite{wang2015vision} and classify abnormal sound sources. For example, one study trained a convolutional neural network (CNN) model \cite{yang2022overcoming} for detecting and localizing \cite{esfahani2019orinet} salient sound sources by combining audio and visual features \cite{masuyama2020self,jain2021vinet}. Several other studies \cite{zhao2022visually,jiang2022egocentric,berghi2021visually, qian2021multi,he2018deep,li2022multi,yun2021pano} have also focused on localizing speakers in images based on 3D sound to improve target detection accuracy. Additionally, incorporating visual depth and infrared modalities to train audio networks for detecting moving vehicles has been demonstrated in a recent study \cite{valverde2021there,gan2019self}. These studies highlight the potential of audio-based approaches for salient sound source detection and target localization \cite{nguyen2023vr,esfahani2018new,nguyen2021liro,esfahani2020local,nguyen2021miliom,esfahani2019deepdsair}.

Previous research has primarily focused on recognizing active speakers or detecting loud sounds, such as vehicles. However, this assumption is not applicable to real-life scenarios where people aren't always talking or driving cars in office hallways. Instead, humans can correlate visual input with distinct footstep sounds while walking to avoid collisions with other pedestrians, making audio an affordable and perceptible modality \cite{esfahani2020unsupervised}. Integrating audio and vision can provide unique advantages such as improved object detection \cite{yuan2014autonomous}, better occlusion handling \cite{wang2017heterogeneous}, and increased robustness against challenging lighting conditions and noise. However, before incorporating audio into robotic perception systems, challenges such as the limited sensing range of audio sensors, interference from multiple audio sources, and a lack of standardized approaches for multi-sensor calibration \cite{wang2015automatic} and synchronization must be addressed.

\begin{figure*}[t]
 \centering
  \includegraphics[width=0.9\linewidth]{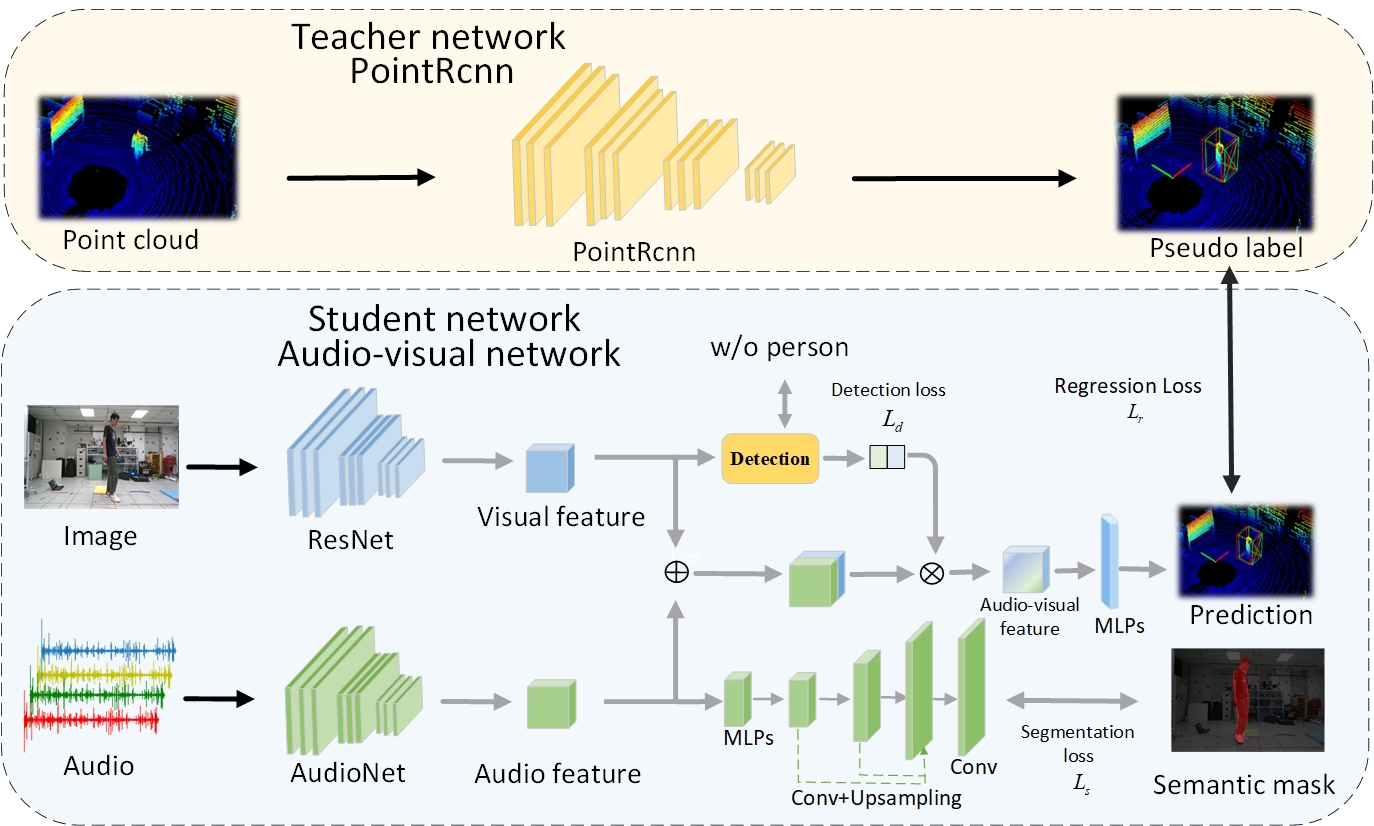}
  \caption{ The audio-visual pedestrian detection network adopts a teacher-student architecture. A well-trained PointRCNN is used to generate 3D bounding box from point cloud data to supervise the learning of audio-visual network. During inference, the network can obtain the 3D bounding box of the pedestrian using only audio and image data.}
   \label{fig:architecture}
\end{figure*}

\section{Proposed Frameworks}
We propose a self-supervised audio-visual network for 360 degree field of view 3D pedestrian detection. To optimize the utilization of audio and visual data, we have incorporated an attention mechanism that assigns weights based on the surrounding environment. Additionally, we have implemented a multi-task learning structure in the network to further enhance its performance. 
We will begin by presenting the core idea of our proposed model and its problem formulation. Next, we will provide a detailed explanation of the network structure. Following that, we will introduce a feature fusion method. Finally, we will elaborate on an auxiliary task that we have incorporated into the model.
\subsection{Cross-modal Self-supervised Learning}
The network is designed to learn in a self-supervised manner by exploiting the natural correspondence between location, image, and audio, without the need for labeled data. Specifically, during the training process, the point cloud data $P$ is fed into a well-trained 3D object detection network $N_p$ to generate a bounding box $Y$. This pseudo label is then utilized to supervise the learning of the audio-visual network $N_{av}$. The audio-visual network takes in both image $I$ and microphone array data $A_n$ as input and extracts complementary features from these two modalities, where $n$ denotes the $n^{th}$ channel of the microphone array. 

To facilitate knowledge transfer from the 3D object detection network to the audio-visual network, we transform the object detection task into a regression problem. The obtained audio-visual features $F_{av}$ are fed into multiple fully connected layers, which are trained to regress the output $\hat{Y}$ close to the pseudo 3D detection boxes $Y$. In this work, Mean Squared Error (MSE) loss is adopted as the regression loss $L_r$ for the object detection task:
\begin{equation}
    {L}_{r} = (Y-\hat{Y})^2.
\label{equation:loss}
\end{equation}
During inference, the network can obtain the 3D bounding box of the pedestrian using only the image and audio inputs, without requiring any LIDAR point cloud data.
\subsection{Forward Audio-visual Pedestrian Awareness Network}
The network structure proposed in this work is based on the teacher-student architecture, as illustrated in Fig. 2. For the purpose of generating pseudo labels, we chose PointRcnn\cite{shi2019pointrcnn} as the teacher network. PointRcnn is a 3D detection model that operates on point cloud data, and it comprises two main components. First, it segments foreground points and generates 3D proposal boxes. Then, based on the generated proposals, it adjusts and refines the detection boxes for better accuracy. PointRcnn has the capacity to  efficiently process sparse point cloud data and has shown its effectiveness on various datasets.

The student network consists of two components: a visual line-of-sight inference network and an audio-based spatial-temporal echo localization network. For the choice of visual network, we utilize a ResNet50 \cite{he2016deep} pretrained on imagenet \cite{russakovsky2015imagenet} for image feature extraction. ResNet has been widely utilized as the backbone for various tasks such as image classification and semantic segmentation. To minimize computational overhead, ResNet is frozen during training, and an additional convolutional layer is appended after ResNet to learn image features.

The audio network is designed to capture the positional information of the sound source. To accomplish this, each 0.4-second audio segment is converted into a mel-spectrogram. Next, the spectrograms from four microphones are stacked along the channel dimension and fed into the network as the input. To extract pedestrian location information from the spectrogram, the audio network utilizes two types of convolutional kernels, as depicted in Fig. 3. The first is a set of time convolution kernels with the same width as the spectrogram sweeps from left to right. These kernels aim to extract time arrival differences information. Then, a batch of frequency convolution kernels with the same length as the spectrogram scan from top to bottom to extract the intensity information for some certain frequency bands of different spectrograms at the same time. It is worth noting that we adopt a multiscale feature extraction strategy, where different sizes of time and frequency convolution kernels are used for feature extraction to obtain more comprehensive information. Then, the features from different scales are concatenated along the dimension to obtain the time and frequency features. Finally, the time and frequency features are concatenated and passed through a fully connected layer to obtain the final audio feature.

\subsection{Audio-visual Feature Fusion}
Commonly used multimodal feature fusion methods include concatenation, addition, and multiplication, etc. Nonetheless, when pedestrians are positioned outside the camera's field of view, the visual characteristics do not encompass a significant amount of helpful information. Directly using the above feature fusion methods, such as concatenating audio and visual features could confuse the network when trained on datasets containing pedestrians both within and outside the camera's field of view.

In order to make better use of the synergies between audio and visual modalities, we propose an attention mechanism that assigns weights to the fusion of audio and visual features, depending on the input visual data. Specifically, we merge the audio and visual features along the channel dimension to form the preliminary audio-visual feature. We then attach a detection head to the visual features to identify pedestrians in the image. The output $\hat{D}$ is a confidence score indicating the likelihood of the presence of pedestrians in the FOV. This confidence score is then multiplied with the preliminary audio-visual feature to achieve weights assignment. Finally, we sum the fused features along the channel dimension to obtain the final audio-visual feature. Notably, the detection is achieved in a supervised way but also without any manually labeled data. We utilize a well-established Yolov4 \cite{bochkovskiy2020yolov4} pretrained on Pascal VOC 2012 dataset \cite{pascal-voc-2012} for human detection, and the resulting detection output $D$ is utilized as the pseudo label to supervise the detection task. The detection loss $L_d$ is defined by:

\begin{equation}
    {L}_{d} =-\frac{1}{N}\sum_{i=1}^{N} D_i\log(\hat{D}_i) + (1-D_i)\log(1-\hat{D}_i).
\label{equation:detection_loss}
\end{equation}

\begin{figure*}[t]
  \centering
   \includegraphics[width=1\linewidth]{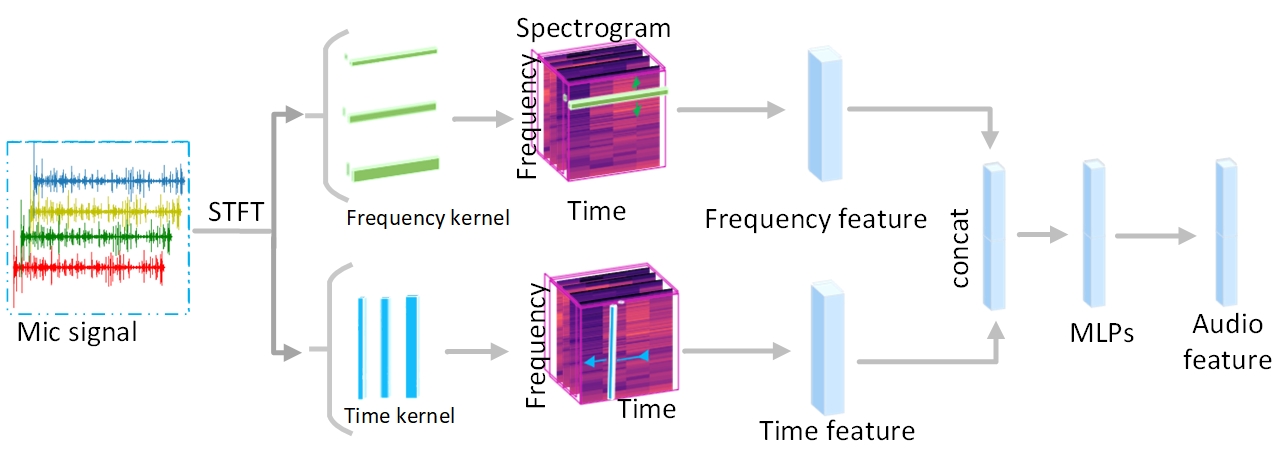}
   \caption{The Structure of Audionet. Two types of convolutional kernels are used to extract time and frequency features from the input spectrogram}
   \label{fig1}
\end{figure*}

\subsection{Multi-task Learning}
To enhance the accuracy and robustness of the audio features in localizing pedestrians, we propose an auxiliary task for the audio network - predicting semantic labels of pedestrians within the FOV.
To implement this, we utilize a PSPNet model \cite{zhao2017pyramid} pre-trained on VOC 2012 dataset to obtain pseudo labels $S$ to supervise the audio network for semantic segmentation. The extracted audio features are passed through a fully connected layer and resized to a 2D vector of (32, 32, 1).  Subsequently, the vector is restored to the original image size using three convolutional layers and upsampling layers. The obtained features from the convolutional layers are stacked and passed through a 1x1 convolutional layer with 2 channels and a softmax activation function to obtain the segmentation result $\hat{S}$. The segmentation loss $L_s$, which is cross-entropy, is employed to supervise the segmentation subnetwork:

\begin{equation}
    {L}_{s} =-\frac{1}{N}\sum_{i=1}^{N} S_i\log(\hat{S}_i) + (1-S_i)\log(1-\hat{S}_i).
\label{equation:segmentationn_loss}
\end{equation}
With the segmentation loss, the total loss $L_t$ is defined by:

\begin{equation}
    {L}_{t} = L_r + \lambda_1 L_d+\lambda_2 L_s ,
\label{equation:total_loss}
\end{equation}  
where $\lambda_1$ $\lambda_2$ are two hyperparameters to adjust the weight between different losses. 




\section{Experiment and Evaluation}
\subsection{Dataset}
\begin{figure}[t]
  \centering
   \includegraphics[width=0.98\linewidth]{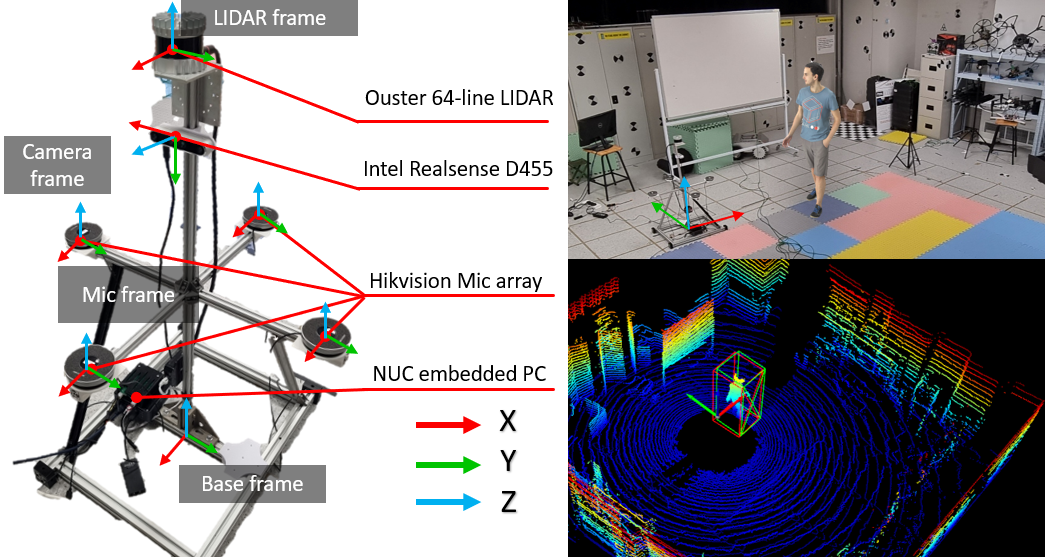}
   \caption{The multi-modality data collection suit used in our experiment.}
   \label{firig1}
\end{figure}

Due to the lack of a publicly available dataset that includes 3D sound, point cloud, and image data, we created a new multimodal pedestrian dataset using a custom rig shown in Fig. \ref{firig1}. The rig comprises an Intel Realsense D455 camera, an omnidirectional Hikvision microphone array, and an Ouster 64-line LIDAR. The microphone array includes four microphones positioned in a cross pattern, with each microphone located 50cm away from the center point, and records the pedestrian's sound at a sampling rate of 48kHz with a sample format of S16LE producing rosbag frequency of 42Hz. The Realsense D455 camera with a wide-angle lens captures images of pedestrians in the field of view at a 20 fps frame rate, with a pixel resolution of 1280$\times$720. Moreover, an Ouster 3D-LIDAR, mounted atop the rig, captures pedestrian point cloud data at a frequency of 10Hz with a vertical field of view of 45\textdegree. We collected data from nine different individuals within a 4.5m$\times$3.5m$\times$3m area, with each person randomly walking around for 5-6 minutes. The rig is positioned at the center of the area to record data from both within and outside the field of view. The data was captured into a NUC11PAHi7000 quad-core i7 embedded PC equipped with 2TB SSD. The data from each sensor are recorded using the Robot Operating System (ROS) into a rosbag and synchronized based on the timestamps provided by ROS.





\begin{figure*}[t]
  \centering
   \includegraphics[width=0.98\linewidth]{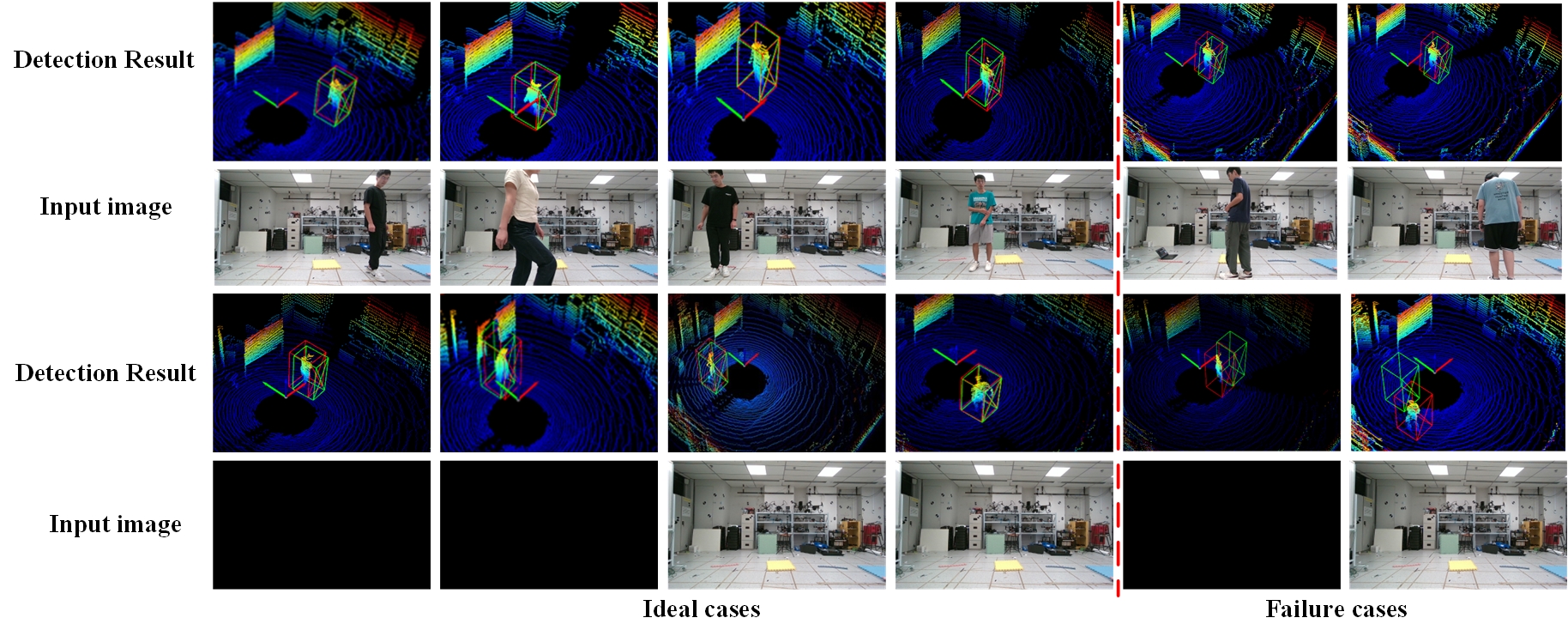}
   \caption{The visualization results of the proposed method. The first row presents the detection results in the case of the pedestrian in line of sight of the camera, while the third row shows the results in the dark environment and when the pedestrian is out of view. The red bounding box is the ground truth, while the green box is the predicted result.}
   \label{figresult}
\end{figure*}

\subsection{Experimental Setting}
\textbf{Data preparation}: To prepare the data for training, we divided all audio signals into small pieces of 0.4 seconds. From each audio segment, we extracted the image and point cloud data at its central moment to create a sample. The processed samples were then divided into two parts: 80\% for training and 20\% for validation. For a more accurate evaluation of the validation set, we manually annotated its data. It's important to note that the training data does not include any labels and the network learning was conducted entirely in a self-supervised mode. This allows us to assess the performance of different methods on the validation set.

To account for the possibility of silent audio segments, we excluded the audio segments with energy lower than the average energy. Subsequently, the remaining audio segments were transformed into mel-spectrograms with a window size of 2048 and a hop size of 1024. We resized the obtained spectrogram and the RGB images into the 256$\times$256 for network input.

\textbf{Implementation Detail}: To train our models, we utilized an Adam optimizer with a batch size of 16 and an initial learning rate of 1$\times$e-4 for 50 epochs. All the experiments are conducted on a GeForce GTX 1080Ti GPU. The hyperparameters $\lambda_1$ and $\lambda_2$ are both set to 0.3. To enhance the network's ability to locate objects in challenging visual environments, we applied various levels of brightness reduction to the images during training, using randomization for data augmentation.

\textbf{Evaluation Metrics}: We employed the average precision (AP) as the evaluation metric for our methods. The performance of the model is tested with AP at an intersection over union (IOU) ranging from 0.2 to 0.5, with an interval of 0.05. To measure the distance between the center of the predicted bounding box and the ground truth, we used the center point distance matrix \cite{gan2019self}. For pedestrian detection, the spatial position ($P_x$,$P_y$) of the person is essential. In this study, we computed the center distance $CD_x$ and $CD_y$ using the following formula: $D_x=\frac{1}{N}\sum_{n=1}^{N}(P_{x}^{n}-G_{x}^{n})$ and $D_y=\frac{1}{N}\sum_{n=1}^{N}(P_{y}^{n}-G_{y}^{n})$, where $G_x$ and $G_y$ represent the center of the ground truth and $N$ denotes the total number of samples.



\subsection{Comparison Experiment}
We compare the proposed method with several baseline networks:

\textbf{VisualNet}: We trained a network \cite{he2016deep} that solely takes images as input and then utilizes a visual network for feature extraction. The visual network used in this approach is identical to the proposed model, while the parameters for the ResNet are trainable.

\textbf{AudioNet}: The audio network utilized the proposed audio feature extraction network developed in this paper for pedestrian detection using only audio data. However, the semantic segmentation branch was not applied during the training stage.

\textbf{SpectrogramNet}: We utilized one of the microphone signals and converted it to a spectrogram to use as input for audio feature extraction. Similar to the Audio Network, the proposed Spectrogram Network did not utilize the auxiliary task. 

\textbf{VSSDL}: We chose to use VSSDL \cite{berghi2021visually}, a visual supervised audio feature extraction network, as our baseline. This network is employed to extract audio features, and the auxiliary task is implemented in this approach.

\begin{table}[]
\caption{The comparison results in the line of sight and under ideal illumination. Lower deviation $D_x$, $D_y$ represents better results, while higher AP represents better results.}
\centering
\renewcommand\arraystretch{1.25}
\begin{tabular}{c|c|c|c|c}
\hline
Approach       & $D_x$& $D_y$ & AP@0.3  &AP@Ave  \\ \hline
VisualNet \cite{he2016deep} &   0.42     &   0.55  &  38.6\%    &   37.1\%  \\ 
AudioNet  &   0.58 &    0.6 &   29.0\%   &   24.8\%   \\ 
SpectrogramNet           &   0.49   &  0.64      &38.0\% &  33.5\%    \\ 
VSSDL \cite{berghi2021visually}      & 0.62       &  0.68      & 28.0\%     & 24.8\%     \\ \hline
AV-PedAware(Ours)           &  \textbf{0.40}     &   \textbf{0.41}     &   \textbf{49.8\%}   & \textbf{42.7\%}     \\ \hline
\end{tabular}

\label{tabel:result}
\end{table}

The experimental results conducted under normal illumination conditions are shown in Table \ref{tabel:result}. It can be observed that by combining visual and auditory modalities, AV-PedAware outperforms VisualNet and AudioNet by an average AP of 5.6\% and 17.9\%, respectively. These results indicate that combining information from both modalities can enhance detection accuracy by capitalizing on the strengths of each modality.

A comparison of the proposed model's performance with SpectrogramNet was carried out, and as expected, SpectrogramNet achieved an average AP of 33.5\%, which was 9.2\% lower than that of AV-PedAware. To overcome the limitation of location information contained in a single microphone signal, multiple microphones were employed, enabling the network to extract more location features such as differences in arrival times. The network's detection ability could be enhanced further with the availability of more microphones.

The performance of our method was further compared with VSSDL, which employs a deep structure with seven convolutional layers and four fully connected layers for audio feature extraction, achieving excellent results in speaker detection. However, in our task, VSSDL obtains an average AP of 24.8\%, which is 17.9\% lower than that of AV-PedAware. Furthermore, VSSDL exhibits the highest center distance values, with $D_x$ and $D_y$ values of 0.62 and 0.68, respectively, among all methods. This is likely due to the presence of noise interference in the audio collection and the relatively simplistic features extracted from the 0.4s audio segment. Consequently, using a deeper network for feature extraction may lead to overfitting and negatively impact detection accuracy. Conversely, AudioNet, developed for our task, aims to extract more robust features such as time arrival difference and energy difference from audio signals, resulting in more precise detection outcomes.

\begin{table}[]
\centering
\renewcommand\arraystretch{1.25}
\caption{The comparison results under dark environment.}
\begin{tabular}{c|c|c|c|c}
\hline
Approach       & $D_x$& $D_y$ & AP@0.3  &AP@Ave  \\ \hline
VisualNet \cite{he2016deep} & 1.63   &   1.49   &  0\%   &   0\%    \\ 
AudioNet  &   0.58 &    0.6 &   \textbf{29.0\%}   &   24.8\%   \\ 
SpectrogramNet & 0.83   &   0.83   &  15.1\%   &   12.9\%   \\ 
VSSDL \cite{berghi2021visually} & 0.84  &   0.84    & 3.4\%   &  2.3\%   \\ \hline
AV-PedAware(Ours)              &   \textbf{0.51}  &  \textbf{ 0.48 }  & 27.0\%  &  \textbf{27.7\%}    \\ \hline
\end{tabular}
\label{tabel:result2}
\end{table}

\subsection{Experiments under dark environment}
To evaluate the performance of the methods under extreme visual conditions, we also test the model in a dark environment setting. 

The experimental results presented in Table \ref{tabel:result2} show that VisualNet fails to perform under harsh visual conditions, as it relies solely on image information for detection. Consequently, the network cannot capture useful information in dark environments. Despite a decrease in performance in poor lighting conditions, our proposed method achieves the highest average AP and the lowest center point distance. It is noteworthy that both AudioNet and AV-PedAware rely on audio data collected by four microphones in dark environments. Nevertheless, the average AP of our method is 2.9\% higher than that of AudioNet. One possible explanation for this is that the auxiliary task incorporated in our method allows the network to learn more position information from audio data.

Fig. \ref{figresult} shows the visualization results of pedestrian detection using our proposed method. The results demonstrate the effectiveness of our method in detecting pedestrians both inside and outside the field of view with ambient machinery noises, as well as in the dark environments. However, as sound signals are susceptible to noise interference, there are instances where the network fails to accurately detect the pedestrian. Addressing this challenge will require further research.

\subsection{Ablation study}
This section analyzes the effectiveness of the attention mechanism and the inclusion of the auxiliary task in the proposed method. Table \ref{tabel:ablation1}  presents the impact of different modules, showing that the attention mechanism contributes the most to the improvement of the model's detection ability. Without incorporating any auxiliary task, the AP with the attention mechanism is 6.8\% higher than the AP without it. Additionally, the inclusion of the semantic segmentation branch enhances the network's performance, resulting in an  improvement in average AP of approximately 1\%.

\begin{table}[]
\centering
\renewcommand\arraystretch{1.25}
\caption{The ablation study of introducing attention and segmentation.}
\begin{tabular}{c|c|c|c|cc}
\hline
Attention       & Segmentation  & $D_x$& $D_y$  &AP@Ave  \\ \hline
$\times$        &$\times$       & 0.67   &   0.74     &   34.8\%    \\ 
$\times$        &$\checkmark$   &   0.45 &    0.46    &   35.5\%   \\ 
$\checkmark$    &$\times$      & 0.41   &   \textbf{0.41}     &   41.6\%   \\ 
$\checkmark$    &$\checkmark$   & \textbf{0.40}  &   \textbf{0.41}     &  \textbf{42.7\%}   \\ \hline

\end{tabular}
\label{tabel:ablation1}
\end{table}




\section{Conclusion}
This paper proposes an innovative audio-visual fusion network for 360-degree pedestrian detection. During the training process, the network is trained in a self-supervised paradigm without any labeled data, leveraging the natural correspondence between different modalities. To improve pedestrian localization in varying visual conditions, an attention mechanism is integrated into the network, allowing for the adjustment of weights between visual and audio cues. Additionally, semantic segmentation is incorporated as an auxiliary task for AudioNet, enabling the network to learn valuable information from the audio input. The proposed method is evaluated on a newly collected dataset comprising multi-modal data for indoor pedestrians, and experimental results demonstrate the method's effectiveness for pedestrian detection tasks. Furthermore, the proposed method can potentially facilitate more advanced robotic tasks.

While existing methods have been successful in detecting single pedestrians in indoor environments, our future work will focus on extending these techniques to the detection of multiple pedestrians in outdoor settings. Moreover, we aim to fully exploit the complementary information provided by audio and visual data to enable the detection of multi-class objects, thus enabling a broader range of applications in robotics.

\bibliographystyle{IEEEtran}
\bibliography{referernces}


\end{document}